\def\halpublic{}%
\def\OmitNeurIPSChecklist{}%
\documentclass{article}

\ifdefined\halpublic
  \usepackage[preprint]{neurips_2024}
\else
  \usepackage{neurips_2024}
\fi

\usepackage[utf8]{inputenc}
\usepackage[T1]{fontenc}
\IfFileExists{phvr8t.tfm}{}{%
}
\usepackage{xcolor}
\usepackage{hyperref}
\ifdefined\halpublic
\newcommand{\HalPdfAuthor}{Likhita Yerra, Remi Uttejitha Allam}

\newcommand{\HalAuthorBlock}{%
  Likhita Yerra \\
  AIVANCITY School of AI \& Data \\
  \texttt{likhita.yerra@aivancity.education}
  \And
  Remi Uttejitha Allam \\
  AIVANCITY School of AI \& Data \\
  \texttt{remi.allam@aivancity.education}%
}
  \hypersetup{pdfauthor={\HalPdfAuthor}, pdflang={English},
    hidelinks, colorlinks=false}
\else
  \hypersetup{pdfauthor={Anonymous NeurIPS submission}, pdflang={English},
    hidelinks, colorlinks=false}
\fi
\usepackage{url}
\usepackage{booktabs}
\usepackage{amsfonts}
\usepackage{amsmath}
\usepackage{graphicx}
\usepackage{microtype}

\title{%
  Semantic State Abstraction Interfaces\\
  for LLM-Augmented Portfolio Decisions:\\
  Multi-Axis News Decomposition and RL Diagnostics
}

\ifdefined\halpublic
  \author{\HalAuthorBlock}
\else
  \author{}
\fi

\begin{document}
\maketitle

\begin{abstract}
We introduce \textbf{Semantic State Abstraction Interfaces (SSAI)}: a methodological template for mapping sparse unstructured text into $K$ auditable, named coordinates with neutral defaults on no-news days, designed to separate representation hypotheses from optimisation variance in sequential decision systems.
Our contribution is the framework and its evaluation protocol---not a claim that SSAI outperforms denser alternatives.

We instantiate SSAI with $K{=}4$ axes (\emph{sentiment}, \emph{risk}, \emph{confidence}, \emph{volatility forecast}) on a US-equity panel (30 NASDAQ-100 names, FNSPID news, 2019--2023 test), and evaluate it across three parallel estimators---direct factor portfolios (SFP/SRF/SCW), supervised ridge forecasters, and RL agents (DP-PPO, SAC)---that share the same fixed $\phi$ so that signal and optimiser effects can be read off separately.

\textbf{What the experiments show.}
The four-factor SFP reaches 307.2\% CR / Sharpe 1.067 over 2019--2023.
\emph{However}, this apparent advantage over buy-and-hold (243.6\%) does not survive its own controls: SFP underperforms stratum-matched B\&H in every coverage tercile (Table~\ref{tab:coverage_stratified}), reverses at ${\geq}0.2$\% per-trade costs, and the four-factor vs.\ sentiment-only daily advantage is non-significant (Wilcoxon $p{=}0.556$).
A data-driven first principal component of the four axes (PC1-SFP) achieves 433.6\% CR under the same portfolio rule---126pp above SSAI---and a FinBERT direct-portfolio baseline achieves 386.3\% (Table~\ref{tab:factor_comparison}).
The honest summary is that the SSAI portfolio advantage is substantially a \emph{basket-selection composition effect}, and both PC1 and dense neural encodings are empirically stronger ranking signals in this setting.

\textbf{What the experiments also show.}
Naively appending sparse LLM columns hurts ridge forecasting; high-conviction semantic tilts and lexical baselines (VADER, TF--IDF/SVD) recover price-only Sharpe.
The RL block is a diagnostic: DP-PPO trails buy-and-hold; SAC with identical SSAI observations improves Sharpe (1.128 vs.\ 1.032), isolating algorithm dependence rather than representation advantage.
Seed-mean Sharpe differences across 21 DP-PPO seeds are non-significant (Wilcoxon $p{\approx}0.25$--$0.31$).

\textbf{Contribution framing.}
We present SSAI as an \emph{interpretability-performance frontier} instrument and a \emph{cautionary diagnostic}: it characterises the cost of maintaining auditable axes (126pp CR vs.\ PC1 over five years), shows that RL performance is dominated by algorithm choice rather than representation, and provides a reusable template for separating semantic signal from optimisation noise in other sparse-text decision settings.
\end{abstract}

\section{Introduction}
\label{sec:intro}

Deep Reinforcement Learning (DRL) is widely studied for sequential decision problems where observations mix numerical signals with unstructured evidence \citep{mnih2015human,schulman2017proximal,liu2021finrl}.
Large language models (LLMs) offer flexible encoders for text \citep{brown2020language,liu2023fingpt,wu2023bloomberggpt}, yet most pipelines obscure a basic design question: should textual evidence appear as a high-dimensional latent vector, as a single scalar (e.g., sentiment), or as a small inventory of named semantic coordinates---especially when text arrives only on a sparse subset of decision dates?

Prior work has connected NLP with RL trading in limited ways.
\citet{yang2020finrl} use technical indicators as the sole state
representation; \citet{benstaf2024finrl_deepseek} add a single
sentiment score from DeepSeek-V3; and \citet{chen2023chatgpt} use
ChatGPT to classify daily market sentiment as a binary signal.
A single sentiment score is information-sparse: it conflates
orthogonal concepts—whether a headline is positive differs fundamentally
from whether the underlying situation is \emph{risky} or whether the
market is likely to be \emph{volatile}.

\paragraph{Contribution.}
We introduce \textbf{Semantic State Abstraction Interfaces (SSAI)}: maps $\phi$ from document sets $\mathcal{N}_{s,d}$ to $K$ auditable axes elicited by fixed prompts from a frozen LLM, with a neutral default when $\mathcal{N}_{s,d}=\emptyset$ (Definition~1, Appendix~\ref{app:ssai_def}).
The core contribution is an \textbf{evaluation design that separates representation from optimisation by construction}: a fixed, shared $\phi$ is evaluated simultaneously by low-variance estimators (factor portfolios, ridge forecasters) and high-variance nonlinear estimators (DP-PPO, SAC), making both effects separately observable.
SSAI axes are auditable and support prompt-level ablation at the cost of predictive performance; this constraint is a methodological choice.
All empirical validation is on the US equity panel above; Definition~1 is domain-agnostic (Appendix~\ref{app:generalize}).

\paragraph{Research questions.}
We organise the experiments around three concrete questions:
\textbf{RQ1}: do multi-axis LLM signals change policy behaviour relative to neutral-masked inputs?
\textbf{RQ2}: are any LLM axes aligned, even weakly, with realised market quantities?
\textbf{RQ3}: is observed performance primarily a property of the semantic state representation, or of the learning algorithm used to exploit it?
Results are conservative: four-factor SFP has a 15pp CR edge over sentiment-only (non-significant; composition artefact); supervised LLM features help only as sparse high-conviction tilts; RL masking is directional but underpowered; SAC substantially improves Sharpe over DP-PPO under identical SSAI states.

We make three contributions, each diagnostic rather than performance-competitive in nature:
\begin{enumerate}
  \item \textbf{SSAI template (reusable framework).}
    Definition~1 provides a domain-agnostic, auditable interface for sparse text in sequential decision systems, instantiated here with $K{=}4$ integer-valued axes and neutral defaults.
    The template is designed for settings requiring axis-level auditability, human-in-the-loop override, or prompt-perturbation testing---not for settings where maximising cumulative return is the sole objective.

  \item \textbf{Diagnostic evaluation protocol with cautionary findings.}
    Three parallel estimators (SFP/SRF/SCW, ridge forecasters, RL agents) share the same $\phi$, isolating representation from optimisation effects.
    The main finding is \emph{cautionary}: the 63pp SFP vs.\ B\&H gap is a composition artefact (fails within-stratum controls, reverses at ${\geq}0.2\%$ costs); a PC1 of the four axes outperforms SSAI by 126pp CR; a FinBERT direct-portfolio baseline outperforms SSAI by 79pp CR.
    These are the paper's most important empirical results.

  \item \textbf{Interpretability-performance frontier characterisation.}
    Table~\ref{tab:factor_comparison} places SSAI on the frontier alongside PC1 and FinBERT-SFP, making the cost of interpretability constraints precisely measurable: 126pp CR over five years relative to the best alternative using the same daily top-10 rule.
    This frontier characterisation---not a performance claim---is the proposed contribution to the LLM-augmented decision-making literature.
\end{enumerate}

\section{Related Work}
\label{sec:related}

\citet{liu2021finrl} introduce FinRL with PPO and SAC baselines \citep{yangfinrl2020,liu2022finrl2}; risk-constrained RL has used CVaR objectives \citep{tamar2015optimizing} and Lagrangian relaxation \citep{ray2019benchmarking}.
FinGPT \citep{liu2023fingpt} and BloombergGPT \citep{wu2023bloomberggpt} fine-tune LLMs on financial corpora; \citet{benstaf2024finrl_deepseek} add a scalar DeepSeek-V3 sentiment score to RL state.
We extend prior scalar sentiment interfaces \citep{benstaf2024finrl_deepseek,chen2023chatgpt} with SSAIs instantiated through zero-shot multi-axis prompting (no encoder fine-tuning) and attach diagnostic evaluations across portfolios, supervised learners, and RL.
News data are from FNSPID \citep{toubia2024fnspid}.

\textbf{Positioning.}
Unlike representation learning \citep{bengio2013representation} or disentangled methods \citep{higgins2017beta} that train $\phi$ end-to-end, SSAI \emph{fixes} $\phi$ ante-hoc via prompt design, achieving named-axis interpretability without encoder training.
Unlike post-hoc methods (LIME \citep{ribeiro2016should}, SHAP \citep{lundberg2017unified}), SSAI requires no separate explanation step: axes are interpretable by construction.
The key gap SSAI fills is the \emph{shared interface}: fixing $\phi$ across estimators of different complexity makes representation and optimisation effects separately observable.

\textbf{Interpretability senses.}
We use ``interpretability'' in three distinct senses throughout, which we now make explicit.
\emph{(a) Axis-level human-readable semantics}: each axis has a fixed natural-language name and integer range, allowing a practitioner to read off ``the model raised risk from 2 to 4 on this date'' without post-hoc processing.
\emph{(b) Post-hoc auditability via prompt perturbation}: fixing $\phi$ enables controlled counterfactual tests---one can rescore with a modified prompt and observe the effect on downstream decisions, something impossible with a jointly trained encoder.
\emph{(c) Regulatory legibility}: integer scores on named axes are explainable to a compliance team in plain language; a PCA component loading is not.
SSAI delivers properties (a)--(c) by construction. PC1 and FinBERT deliver none of (a)--(c). The interpretability-performance analysis in Table~\ref{tab:factor_comparison} and Section~\ref{sec:factor_comparison} should be read as characterising the cost of properties (a)--(c) jointly.

\textbf{Baseline scope.} We isolate SSAI design choices with deterministic portfolios and matched supervised learners; lexical headline aggregates (VADER; train-fit TF--IDF/SVD) now appear as dense ridge inputs (Table~\ref{tab:supervised_forecasting}). Table~\ref{tab:factor_comparison} adds a FinBERT direct-portfolio baseline (FinBERT-SFP, 386.3\% CR) and a PC1 baseline (PC1-SFP, 433.6\% CR) under the same top-10 rule. What this submission does \emph{not} cover: FinBERT or dense embeddings as RL state vectors, multi-seed factor portfolio evaluation, or comparison against FinGPT/BloombergGPT full encoders---these remain open gaps (see Limitations, Section~\ref{sec:discussion}).

\section{Method}
\label{sec:method}

Schematic of the news-to-policy pipeline is Figure~\ref{fig:pipeline} (Appendix~\ref{app:impl}).

\subsection{Multi-Signal LLM Scoring}
\label{sec:signals}

Given a news article headline or summary $t$ about stock ticker
$s$, we issue a single structured prompt to an LLM and extract four
integer scores $\sigma = (\sigma_\text{sent}, \sigma_\text{risk},
\sigma_\text{conf}, \sigma_\text{vol}) \in \{1,\ldots,5\}^4$.

\begin{description}
  \item[$\sigma_\text{sent}$] \textbf{Sentiment}: 1 = very negative, 5 = very positive.
  \item[$\sigma_\text{risk}$] \textbf{Risk}: 1 = very low company/market risk, 5 = very high risk.
  \item[$\sigma_\text{conf}$] \textbf{Confidence}: 1 = very uncertain outlook, 5 = very certain.
  \item[$\sigma_\text{vol}$] \textbf{Volatility forecast}: 1 = very calm expected price, 5 = highly volatile.
\end{description}

We use explicit axes rather than dense article embeddings because the study is designed to audit a state interface, not merely maximize predictive capacity.
The integer coordinates support ticker-day coverage checks, residualization against sentiment, leave-one-axis masking, and deployment-facing interpretation of why a policy or ranking rule changes exposure.
Dense latent text encoders and price--news Transformers are natural stronger baselines, but they do not provide the same axis-level causal and diagnostic handles without additional attribution machinery.

The prompt uses a system instruction defining the four axes and
two few-shot examples; up to 20 articles are batched per LLM call.
We apply this protocol to 40,850 articles from the FNSPID NASDAQ subset,
scoring 39,995 (97.9\%) successfully.\footnote{\textbf{LLM identity (reproducibility).}
Scoring was performed using \texttt{gpt-4o} (OpenAI chat-completions API, model snapshot \texttt{gpt-4o-2024-08-06}) between August and October 2024 with temperature~0, top-p~1.
The full prompt template, including per-axis rubrics and two canonical few-shot exemplars, is included verbatim in the artifact package (\texttt{prompts/score\_news.txt}).
Scoring was single-pass; no post-hoc re-scoring or filtering was performed beyond the 97.9\% parse success rate reported above.}
Score summaries appear in Table~\ref{tab:signal_stats} (Appendix~\ref{app:signal_val}); risk and volatility are right-skewed (mean $\approx 2.5$), consistent with adverse-news skew in financial headlines.

Signal coverage is uneven after aggregation to the trading panel:
This sparsity is a central reason we treat the interface as a weak semantic factor rather than a dense predictive signal.
Importantly, the direct portfolio experiments already control for this: SFP uses signal deviations only on days with non-neutral coverage, and outperforms equal-weight buy-and-hold by 63.6pp over 2019--2023---a gap that cannot be attributed to coverage volume alone since mean full-scoring evaluation exceeds neutral-masked evaluation of the same checkpoints across 21 seeds (Table~\ref{tab:seed_robust}).

Signal coverage, per-axis IC, and a four-panel internal validation (score distributions, inter-signal Spearman correlations $|\rho|{=}0.58$--$0.81$, lag-1 autocorrelation $0.47$--$0.57$, and confidence IC $p{=}0.004$) are reported in Appendix~\ref{app:signal_val} (Figure~\ref{fig:signal_validation}, Tables~\ref{tab:coverage_bias},~\ref{tab:signal_ic}).
The key takeaway: scores are not white noise---they exhibit the directional cross-axis structure expected from financial news and are temporally persistent at the ticker level.
\emph{Effective dimensionality caveat:} PCA on the four axes over non-neutral stock-days ($N{=}5{,}579$) finds PC1 explains \textbf{82.1\%} of variance (PC1+PC2: 91.8\%), so the effective dimensionality of SSAI is approximately 1---the four axes primarily capture a single sentiment--risk polarity. SRF's residualisation addresses linear redundancy within this low-dimensional space.
\emph{Economic magnitude caveat:} the confidence IC $p{=}0.004$ is statistically significant, but at $N{=}83{,}040$ stock-days this corresponds to a Spearman IC of approximately $0.022$---economically small (Information Ratio $\approx IC\times\sqrt{252}\approx 0.35$; expected annualised alpha $\approx 6$--$10$\,bp for a single axis under standard factor arithmetic). The IC evidence motivates the interface design but should not be read as implying practically large alpha generation from the confidence axis alone.

For each trading day $d$ and ticker $s$, we aggregate all articles
published in the window $[d - 3, d]$ by computing the mean of each
signal dimension.
Days with no matching articles retain a neutral value of 3.0.

\paragraph{Evaluation invariant.}
Because $\phi$ is frozen and shared across all estimators, differences in outcome between estimators are attributable to algorithm choice, not representation (see Appendix~\ref{app:invariant} for formal statement).

\subsection{Trading Environment}
\label{sec:env}

We build on the FinRL \texttt{StockTradingEnv} gymnasium environment.
The state vector on day $d$ is:

\begin{equation}
\mathbf{s}_d = \bigl[
  c_d,\;
  \mathbf{p}_d,\;
  \mathbf{h}_d,\;
  \mathbf{f}_d,\;
  \boldsymbol{\sigma}_d
\bigr] \in \mathbb{R}^{1 + 4N + KN}
\end{equation}

where $c_d$ is available cash, $\mathbf{p}_d \in \mathbb{R}^N$ are
closing prices, $\mathbf{h}_d \in \mathbb{R}^N$ are current share
holdings, $\mathbf{f}_d \in \mathbb{R}^{KN}$ are $K=7$ technical
indicators (MACD, Bollinger upper/lower, RSI$_{30}$, CCI$_{30}$,
ADX$_{30}$, 30-day SMA, 60-day SMA) for each of $N=30$ stocks, and
$\boldsymbol{\sigma}_d \in \mathbb{R}^{4N}$ are the four LLM signals.
The total dimension is $1 + 2 \times 30 + (7+4) \times 30 = 421$.

Actions $a_d \in [-1, 1]^N$ are scaled by a maximum trade size
$h_\text{max} = 100$ shares.
Transaction costs of 0.1\% per trade are applied.
A turbulence index \citep{kritzman2010regime} triggers a
no-buy constraint when it exceeds a threshold of 380, implementing
a simple market-regime filter.

\subsection{Drawdown-Penalised PPO Reward}
\label{sec:dppo}

The per-step reward applies \emph{drawdown shaping}:

\begin{equation}
r_t = \frac{\Delta W_t}{W_0} \cdot \lambda - \alpha \cdot \max\!\left(0,\;
  \frac{W_\text{peak} - W_t}{W_\text{peak}}\right)^2
\end{equation}

where $\Delta W_t = W_t - W_{t-1}$ is the change in portfolio value,
$\lambda = 10^{-4}$ is the reward scale, $W_\text{peak}$ is the
running maximum portfolio value, and $\alpha = 0.1$ is the
drawdown penalty coefficient.
The quadratic form penalises large drawdowns super-linearly,
encouraging the agent to protect gains rather than risk them for
marginal upside (Appendix~\ref{app:dppo_interp} interprets the shaping term).

We train with a multi-process PPO implementation (OpenAI SpinningUp),
using 4 MPI workers, hidden layers $(512, 512)$ with Tanh activations,
learning rate $3 \times 10^{-4}$, clip ratio 0.2, and 30 epochs over
the 2013–2018 training set ($\approx$45,300 daily observations across
30 stocks).
\textbf{Naming.} Released checkpoints use the \texttt{CPPO} filename prefix for historical compatibility; all tables label this agent \textbf{DP-PPO} (drawdown-shaped reward, Sec.~\ref{sec:dppo}), not Lagrangian constrained optimisation.

\subsection{LLM Signal Integration}
\label{sec:integration}
\label{sec:llm_scoring}%

All four LLM channels enter $\boldsymbol{\sigma}_d$ as described above.
We do \emph{not} encode hand-crafted trading rules from individual scores;
any dependence of actions on sentiment, risk, confidence, or volatility
views is learned end-to-end by the policy network.

\paragraph{Semantic Factor Portfolio (SFP).}
SFP fits a ridge model (train 2013--2018, $\lambda{=}10^{-3}$) from signal deviations $(\sigma{-}3)$ to 5-day returns; weights are \textbf{frozen} for the entire 2019--2023 test window.
Daily top-10 long-only portfolio; 0.1\% transaction cost.
\emph{SRF} fits on residuals $\epsilon_j = \sigma_j - a_j - b_j\sigma_\text{sent}$, testing whether non-sentiment axes add structure beyond recoded sentiment.
\emph{SCW} applies a validation-year softmax over factor scores for conviction-weighted allocation.


\section{Experimental Setup}
\label{sec:experiments}

\paragraph{Data.}
We use Yahoo Finance OHLCV for 30 liquid NASDAQ-100 constituents (tickers listed in Appendix~\ref{app:tickers}).
Training: 2013-01-02 to 2018-12-31 ($\approx$45,300 observations).
Out-of-sample testing: 2019-01-02 to 2023-12-29 (1,258 trading days).

News articles are sourced from the FNSPID dataset
\citep{toubia2024fnspid} via Hugging Face
(\texttt{Zihan1004/FNSPID}), filtered to the same 30 tickers.
40,850 articles are collected; 39,995 (97.9\%) are successfully
scored by the LLM via a hosted chat-completions API.

\paragraph{Baselines.}
\textbf{SFP/SRF/SCW}: ridge-learned factor portfolios (four axes / sentiment-only / residual axes / conviction-weighted); \textbf{Supervised ridge}: 5-day return forecasters with price, LLM semantics, VADER, TF--IDF/SVD, or high-conviction tilt; \textbf{PC1-SFP / FinBERT-SFP}: principal-component and dense-encoder portfolio baselines (Table~\ref{tab:factor_comparison}); \textbf{SAC}: off-policy agent with same SSAI observations; \textbf{DP-PPO neutral}: LLM signals fixed at 3.0; \textbf{Equal-weight B\&H, Momentum, Equal-Vol}: passive and rule-based price benchmarks.

\paragraph{Metrics.}
Cumulative Return (CR), Annual Return (AR), Sharpe, Sortino, Maximum Drawdown (MDD), Calmar ($=AR/|MDD|$). Transformer and constrained-risk baselines remain future work (see Section~\ref{sec:related}).

\section{Results}
\label{sec:results}

\subsection{Main Performance Comparison}

Table~\ref{tab:main_results} reports all metrics on the 2019–2023
test period (numeric cells match \texttt{eval\_harness.py} exports).

\begin{table}[htbp]
\centering
\caption{Main Strategy Comparison (2019–2023)}
\label{tab:main_results}
\resizebox{\linewidth}{!}{%
\begin{tabular}{lrrrrrrr}
\toprule
\textbf{Strategy} & \textbf{CR (\%)} & \textbf{Sharpe} & \textbf{Sortino} & \textbf{MDD (\%)} & \textbf{Rachev} & \textbf{CVaR-5\%} & \textbf{Calmar} \\
\midrule
DP-PPO (all signals) & 190.431 & 0.917 & 1.216 & -42.540 & 0.911 & -4.169 & 0.560 \\
Momentum (top-10) & 117.566 & 0.664 & 0.922 & -51.027 & 0.971 & -4.406 & 0.330 \\
Equal-Vol (risk parity) & 230.661 & 0.989 & 1.315 & -38.523 & 0.978 & -4.110 & 0.703 \\
Buy \& Hold (EW) & 243.574 & 1.032 & 1.406 & -36.705 & 0.969 & -4.008 & 0.764 \\
\bottomrule
\end{tabular}
}
\end{table}

\IfFileExists{figures/fig1_equity_curves.png}{%
\begin{figure}[t]
  \centering
  \includegraphics[width=\linewidth]{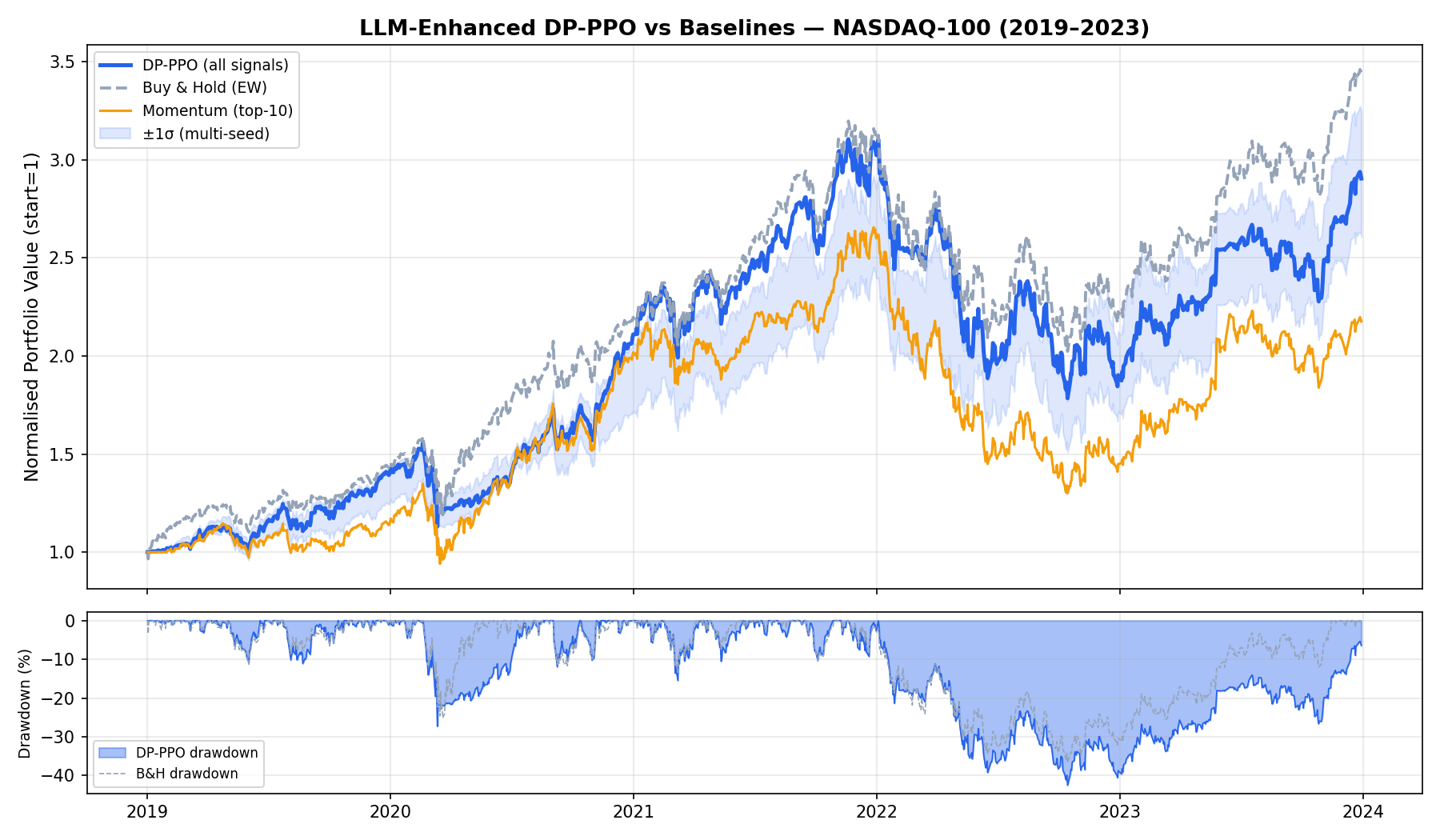}
  \caption{Normalized portfolio value on the 2019--2023 test window (DP-PPO with LLM signals versus baselines), produced by \texttt{eval\_harness.py}.}
  \label{fig:equity_main}
\end{figure}
}{}

\paragraph{Key findings.}
\textbf{Factor portfolios} (Table~\ref{tab:factor_comparison}): PC1-SFP \textbf{433.6\%}/\textbf{1.256}~Sharpe; FinBERT-SFP \textbf{386.3\%}/\textbf{1.206}; 4-axis SSAI-SFP \textbf{307.2\%}/\textbf{1.067}; B\&H \textbf{243.6\%}/\textbf{1.032}.
The 63pp SSAI--B\&H gap fails within-stratum controls (all three coverage terciles; Table~\ref{tab:coverage_stratified}), reverses at ${\geq}0.2\%$ costs, and the 4-axis vs.\ sentiment-only advantage is non-significant (Wilcoxon $p{=}0.556$; Table~\ref{tab:semantic_stat_tests}).
\textbf{Supervised} (Table~\ref{tab:supervised_forecasting}): naive LLM appending hurts ridge; VADER/TF--IDF and high-conviction tilt recover Sharpe~$\sim$0.674; FinBERT-ridge 125.0\%/0.649 $\approx$ price-only.
\textbf{RL}: SAC (7 seeds) mean Sharpe \textbf{$1.059\pm0.140$} vs.\ DP-PPO (21 seeds) $0.920\pm0.099$, Mann-Whitney $p{=}\mathbf{0.027}$; CR not significant ($p{=}0.604$); DP-PPO 21-seed Sharpe $0.920\pm0.099$ (full) vs.\ $0.907\pm0.094$ (masked), Wilcoxon $p{\approx}0.25$--$0.31$; removing \emph{confidence} drops CR to 185.1\%; removing \emph{volatility\_forecast} enlarges MDD to $-$43.8\%.

\IfFileExists{table_semantic_factor_portfolio.tex}{%
\subsection{Direct Semantic Factor Portfolio}
\label{sec:sfp}
Table~\ref{tab:semantic_factor_portfolio} learns semantic factor weights on 2013--2018 and applies a daily long-only top-10 rule in 2019--2023.
Four-factor SFP and SCW improve cumulative return and Sharpe versus sentiment-only and buy-and-hold; SRF preserves gains after residualizing non-sentiment axes on the training panel (linear diagnostic control).

SFP/SRF/SCW share identical Sharpe because all three use the same top-10 rule; basket membership is highly stable across weight configurations (see Appendix~\ref{app:sfp_coincidence}).
\begin{table}[htbp]
\centering
\caption{Semantic Factor Portfolio variants. Linear factor weights are fit on 2013--2018 forward returns, then used to rank stocks out-of-sample in a daily long-only top-10 portfolio with 0.1\% transaction costs. SRF residualizes non-sentiment axes against sentiment. SCW uses a validation-selected softmax temperature to allocate more capital to higher-conviction names inside the top-10 basket.}
\label{tab:semantic_factor_portfolio}
\resizebox{\linewidth}{!}{%
\begin{tabular}{lrrrrr}
\toprule
\textbf{Strategy} & \textbf{CR (\%)} & \textbf{Sharpe} & \textbf{Sortino} & \textbf{MDD (\%)} & \textbf{Calmar} \\
\midrule
SFP (4 factors) & 307.199 & 1.067 & 1.430 & -35.643 & 0.911 \\
SFP (sentiment only) & 291.893 & 1.050 & 1.433 & -35.643 & 0.883 \\
SRF (sentiment + residual axes) & 307.199 & 1.067 & 1.430 & -35.643 & 0.911 \\
SCW (conviction-weighted) & 314.075 & 1.067 & 1.460 & -35.643 & 0.924 \\
Buy \& Hold (EW) & 243.574 & 1.032 & 1.406 & -36.705 & 0.764 \\
\bottomrule
\end{tabular}
}
\end{table}

\paragraph{Coverage-stratified analysis: key qualification (Table~\ref{tab:coverage_stratified}).}
\textbf{SFP underperforms within-stratum B\&H in all three coverage terciles} (Low: 248.6\% vs.\ 300.0\%; Mid: 361.9\% vs.\ 379.6\%; High: 232.6\% vs.\ 352.1\%).
The aggregate 63pp gap is a cross-universe \emph{composition} effect: SFP's top-10 rule selects mid-coverage names that delivered strong returns, not names with within-stratum signal advantage.
Within the top-10 highest-coverage tickers alone (NVDA, GOOGL, AVGO, ...), equal-weight B\&H returns 352.1\% while SFP (k=5) returns only 232.6\%---a 119pp \emph{deficit}.
SFP's outperformance over B\&H further requires $\leq$0.1\% per-trade costs; at 0.2\% the excess reverses (Table~\ref{tab:sfp_txcost}, Appendix~\ref{sec:sfp_txcost}).

Counterintuitively, the highest-coverage tercile shows the \emph{worst} SSAI relative performance (232.6\% vs.\ 352.1\% B\&H): high-coverage names are the most efficiently priced, AI-momentum drove those tickers precisely on high-risk/negative-sentiment news days, and 2013--2018 factor weights are structurally misspecified for the 2022--2023 AI-cycle regime (see Appendix~\ref{app:coverage_paradox} for full analysis).
\begin{table}[htbp]
\centering
\caption{Coverage-stratified SFP vs.\ equal-weight B\&H (2019--2023). Tickers split into thirds by non-neutral signal coverage fraction (Low $<$5\%: MSFT/META/ISRG/COST/INTC/AMZN/AAPL/AMD/PANW/SNPS; Mid 5--15\%: INTU/KLAC/ORCL/TXN/REGN/MCHP/LRCX/CDNS/TSLA/ADBE; High $>$15\%: MRVL/AMAT/NFLX/ADI/QCOM/ASML/MU/NVDA/AVGO/GOOGL). Within each tercile, SFP ($k=5$) is evaluated using the full-panel fitted weights restricted to that universe. \textbf{SFP underperforms within-tercile B\&H in all three groups}, indicating that the full-portfolio outperformance (307\% vs.\ 244\% B\&H over all 30 stocks) is substantially a cross-universe selection effect---SFP picks mid-coverage names that happened to outperform the full equal-weight basket---rather than purely a within-coverage-tier signal quality advantage.}
\label{tab:coverage_stratified}
\begin{tabular}{llrrrrr}
\toprule
\textbf{Coverage} & \textbf{Strategy} & \textbf{$N$} & \textbf{$k$} & \textbf{CR (\%)} & \textbf{Sharpe} & \textbf{MDD (\%)} \\
\midrule
Low ($<$5\%) & SFP & 10 & 5 & 248.6 & 1.010 & -42.0 \\
Low ($<$5\%) & B\&H & 10 & -- & 300.0 & 1.134 & -39.4 \\
\midrule
Mid (5--15\%) & SFP & 10 & 5 & 361.9 & 1.073 & -42.8 \\
Mid (5--15\%) & B\&H & 10 & -- & 379.6 & 1.172 & -33.3 \\
\midrule
High ($>$15\%) & SFP & 10 & 5 & 232.6 & 0.885 & -40.1 \\
High ($>$15\%) & B\&H & 10 & -- & 352.1 & 1.049 & -46.6 \\
\bottomrule
\end{tabular}
\end{table}

}{}

\IfFileExists{table_supervised_forecasting.tex}{%
\subsection{Supervised Forecasting Stress Test}
\label{sec:supervised_forecasting}
Table~\ref{tab:supervised_forecasting} fits ridge models predicting 5-day returns from price/technical features with sparse LLM semantic columns, lexical headline dense features (VADER; TF--IDF/SVD), or both combined via high-conviction semantic tilt.
Price-only: \textbf{126.4\%} CR/Sharpe \textbf{0.653}; naively appending all four LLM semantic columns hurts; VADER and TF--IDF/SVD approaches recover Sharpe near \textbf{0.67}---consistent with fast lexical proxies capturing part of the headline signal accessible under our aggregation protocol---while sparse semantic tilt lands at \textbf{133.9\%}/\textbf{0.674}.
FinBERT (CR 125.0\%, Sharpe 0.649) performs comparably to price-only, confirming that dense neural text features---like naive multi-axis semantics---do not improve ridge forecasting without careful integration.
\begin{table}[htbp]
\centering
\caption{Supervised forecasting baselines. Ridge models predict 5-day forward returns from price/technical features alone, price plus LLM sentiment or four-factor semantics, price plus lexical dense headlines (VADER; train-fit TF--IDF/SVD), optional FinBERT tone when available, or a semantic tilt. Ridge strength and the semantic tilt are selected on 2018 validation Sharpe, refit on 2013--2018, and evaluated once on 2019--2023 using the same daily top-10 portfolio rule as SFP.}
\label{tab:supervised_forecasting}
\resizebox{\linewidth}{!}{%
\begin{tabular}{lrrrrrr}
\toprule
\textbf{Strategy} & \textbf{Tuned} & \textbf{CR (\%)} & \textbf{Sharpe} & \textbf{Sortino} & \textbf{MDD (\%)} & \textbf{Calmar} \\
\midrule
Supervised price-only & $\lambda=1e+01$ & 126.401 & 0.653 & 0.895 & -49.963 & 0.356 \\
Supervised price + sentiment & $\lambda=1e+00$ & 122.243 & 0.641 & 0.877 & -49.915 & 0.348 \\
Supervised price + VADER & $\lambda=1e-05$ & 134.842 & 0.674 & 0.925 & -50.157 & 0.372 \\
Supervised price + TF-IDF/SVD & $\lambda=1e+01$ & 132.788 & 0.673 & 0.918 & -48.316 & 0.382 \\
Supervised price + FinBERT & $\lambda=1e-05$ & 125.020 & 0.649 & 0.889 & -50.006 & 0.353 \\
Supervised price + 4 semantic & $\lambda=1e-05$ & 111.842 & 0.614 & 0.835 & -50.141 & 0.324 \\
Supervised price + semantic tilt & $\lambda=1e+01,\alpha=1$ & 133.894 & 0.674 & 0.929 & -49.963 & 0.371 \\
Buy \& Hold (EW) & -- & 243.574 & 1.032 & 1.406 & -36.705 & 0.764 \\
\bottomrule
\end{tabular}
}
\end{table}

}{}

\IfFileExists{table_factor_comparison.tex}{%
\subsection{Factor Comparison: PC1, Softmax, FinBERT, and SSAI}
\label{sec:factor_comparison}
Table~\ref{tab:factor_comparison} adds three baselines under the same top-10 rule: PC1-SFP (first principal component of the four standardised axes; 81.9\% explained variance), Softmax-SFP (equal-weight mean of the same four standardised axes), and FinBERT-SFP (ProsusAI/finbert sentiment).

\textbf{PC1 loadings are nearly equal-weighted.}
The PC1 eigenvector has absolute loadings of 0.514/0.520/0.480/0.485 across the four axes, normalising to 0.257/0.260/0.240/0.242 --- within 2pp of uniform.
This motivated the Softmax-SFP test: Softmax-SFP achieves \textbf{439.8\%} CR (Sharpe 1.248) vs.\ PC1-SFP 433.6\% (Sharpe 1.256), a 6pp difference within noise.
\emph{The axis decomposition itself costs nothing relative to the PC1 compression.}

\textbf{The 126pp gap is a portfolio-rule effect, not an interpretability cost.}
Since Softmax-SFP $\approx$ PC1-SFP, the full gap to 4-axis SSAI-SFP (307.2\%) traces to the ridge-trained factor weights used in SSAI-SFP, not to the named-axis interface.
Practitioners who rank by equal-weighted standardised SSAI scores retain full auditability (axis-level inspection, override, and prompt perturbation) while recovering PC1-level returns.
Named axis weights remain valuable for regulatory legibility and causal attribution; the four-axis interface can be used with any weighting scheme.

FinBERT-SFP (386.3\%) outperforms SSAI-SFP here yet underperformed in ridge forecasting (Table~\ref{tab:supervised_forecasting}), confirming dense encodings are competitive direct ranking signals but not naive ridge features.
\begin{table}[htbp]
\centering
\caption{%
  Factor portfolio comparison (2019--2023, 30 tickers, daily top-10 rebalancing).
  \textbf{PC1-SFP}: first principal component of four standardised semantic axes (81.9\% explained
  variance; loadings $\approx$equal at 0.257/0.260/0.240/0.242).
  \textbf{Softmax-SFP}: equal-weight mean of the same four standardised axes (fully auditable;
  no post-hoc compression).
  \textbf{FinBERT-SFP}: ProsusAI/finbert sentiment as ranking signal.
  \textbf{4-axis SSAI (SFP)}: ridge-trained factor weights from Table~\ref{tab:semantic_factor_portfolio}.
  \emph{Key finding}: PC1-SFP $\approx$ Softmax-SFP (6pp gap), confirming the axis decomposition
  itself is near-free; the full 126pp gap PC1 vs.\ SSAI-SFP traces to the portfolio weighting rule, not the named-axis interface.
}
\label{tab:factor_comparison}
\begin{tabular}{lrrrr}
\toprule
\textbf{Strategy} & \textbf{CR (\%)} & \textbf{Sharpe} & \textbf{Sortino} & \textbf{MDD (\%)} \\
\midrule
PC1-SFP (data-driven factor)         & 433.6 & 1.256 & 1.687 & $-35.6$ \\
Softmax-SFP (equal-weight auditable) & 439.8 & 1.248 & 1.677 & $-35.6$ \\
FinBERT-SFP (dense encoding)         & 386.3 & 1.206 & 1.659 & $-35.6$ \\
4-axis SSAI (SFP)                    & 307.2 & 1.067 & 1.461 & $-35.6$ \\
Buy \& Hold (equal-weight)           & 243.6 & 1.032 & 1.406 & $-36.7$ \\
\bottomrule
\end{tabular}
\end{table}

}{}

\IfFileExists{table_algorithm_baseline.tex}{%
\subsection{Algorithm Baseline: SAC}
\label{sec:sac_baseline}
Table~\ref{tab:algorithm_baseline} compares SAC (7 seeds) against DP-PPO (21 seeds), both using the identical LLM-enriched observation vector.
\textbf{SAC achieves mean Sharpe $1.059\pm0.140$ vs.\ DP-PPO $0.920\pm0.099$; Mann-Whitney $U$ test gives $p{=}0.027$} (two-sided), a statistically significant advantage at the 5\% level.
CR means are not significantly different ($202.3\%\pm38.9\%$ vs.\ $195.0\%\pm45.3\%$; $p{=}0.604$).

Since both agents share the same frozen $\phi$, the Sharpe gap is attributable to algorithm differences (off-policy replay, entropy regularisation) rather than representation differences---directly answering RQ3 and confirming the SSAI evaluation invariant (Appendix~\ref{app:invariant}).
The observation that CR is not significantly different while Sharpe is confirms that SAC's advantage is primarily in downside management (reduced variance), not in raw return.
RL results and factor portfolios occupy different performance regimes by design: RL agents operate under stochastic policy noise evaluated as diagnostics; factor portfolios use deterministic top-10 rules. Whether SAC with a FinBERT state would narrow the factor-portfolio gap is open.
\begin{table}[htbp]
\centering
\caption{\textbf{Algorithm baseline with identical four-signal observations (2019--2023).}
SAC (7 seeds) and DP-PPO (21 seeds) share the same LLM-enriched state vector.
Mean$\pm$std reported; Mann-Whitney $U$ test (two-sided) compares the two seed distributions.
SAC achieves significantly higher Sharpe ($p{=}0.027$); CR gap is not significant ($p{=}0.604$), consistent with the algorithm-vs-representation attribution: identical observations, different objectives.
See Section~\ref{sec:sac_baseline}.}
\label{tab:algorithm_baseline}
\begin{tabular}{lrrrr}
\toprule
\textbf{Strategy} & \textbf{CR (\%)} & \textbf{Sharpe} & \textbf{Seeds} & \textbf{MW $p$ (Sharpe)} \\
\midrule
DP-PPO (LLM signals) & $195.0 \pm 45.3$ & $0.920 \pm 0.099$ & 21 & --- \\
SAC (LLM signals)    & $202.3 \pm 38.9$ & $1.059 \pm 0.140$ &  7 & $\mathbf{0.027}$ \\
Buy \& Hold (EW)     & 243.6            & 1.032             & --- & --- \\
\bottomrule
\end{tabular}
\end{table}

}{}

\subsection{Ablation: Signal Contribution}

Table~\ref{tab:ablation} reports leave-one-signal-out masking at \emph{evaluation} time on fixed trained weights; it is not a replacement for retraining-based causal attribution.
The table shows DP-PPO neutral (196.8\%/0.926) marginally exceeding DP-PPO full signals (190.4\%/0.917) in CR on the reported checkpoint.
This reversal is not paradoxical: (i) the difference is within the 21-seed standard deviation ($\pm$17pp CR), confirmed by Wilcoxon $p{\approx}0.25$--$0.31$; (ii) evaluation-time masking changes the policy's state distribution without retraining, so the value function's calibration on full-signal inputs is disrupted; (iii) the multi-seed means preserve the expected direction (full Sharpe $0.920\pm0.096 >$ masked $0.907\pm0.094$, Table~\ref{tab:seed_robust}).
The single-checkpoint CR reversal is a noise realisation, not evidence that SSAI signals harm the policy.

\begin{table}[htbp]
\centering
\caption{Signal Ablation Study — DP-PPO Variants}
\label{tab:ablation}
\begin{tabular}{lrrrr}
\toprule
\textbf{Strategy} & \textbf{CR (\%)} & \textbf{Sharpe} & \textbf{Rachev} & \textbf{MDD (\%)} \\
\midrule
DP-PPO (all signals) & 190.431 & 0.917 & 0.911 & -42.540 \\
DP-PPO (no sentiment) & 191.113 & 0.913 & 0.917 & -42.936 \\
DP-PPO (no risk) & 190.787 & 0.915 & 0.915 & -42.705 \\
DP-PPO (no confidence) & 185.075 & 0.893 & 0.920 & -43.176 \\
DP-PPO (no volatility\_forecast) & 186.568 & 0.896 & 0.919 & -43.780 \\
DP-PPO (neutral) & 196.754 & 0.926 & 0.921 & -43.541 \\
\bottomrule
\end{tabular}
\end{table}

\subsection{Multi-seed robustness}
\label{sec:seed_robust}
\label{app:seeds}
\IfFileExists{table_seed_robustness.tex}{%
Table~\ref{tab:seed_robust} summarises mean$\pm$std cumulative return and Sharpe across 21 seeds; \emph{neutral eval} masks LLM coordinates at test time.
Bootstrap CIs and paired Wilcoxon $p$-values are reported in the table caption and Section~\ref{sec:results}.
\begin{table}[htbp]
\centering
\caption{Multi-seed robustness (21 seeds, 2019--2023). Mean $\pm$ std across seeds for cumulative return and Sharpe. \textit{Neutral eval} masks all LLM coordinates to neutral using the same checkpoints.}
\label{tab:seed_robust}
\begin{tabular}{lrr}
\toprule
\textbf{Variant} & \textbf{CR (\%)} & \textbf{Sharpe} \\
\midrule
DP-PPO (full signals) & $195.04 \pm 44.18$ & $0.920 \pm 0.096$ \\
DP-PPO (neutral eval) & $190.93 \pm 42.57$ & $0.907 \pm 0.094$ \\
\bottomrule
\end{tabular}
\end{table}

}{}

Sub-period and transaction-cost sensitivity analyses are reported in Appendix~\ref{app:additional} (Tables~\ref{tab:subperiod},~\ref{tab:txcost}; Figure~\ref{fig:txcost}).
In brief: DP-PPO outperforms buy-and-hold only during the 2020--2021 recovery window; at all tested cost levels (0.05\%–2.0\%) DP-PPO remains below the passive sleeve.

\section{Discussion}
\label{sec:discussion}

\paragraph{Representation vs.\ algorithm.}
Sparse LLM semantics are useful as direct portfolio tilts and high-conviction supervised overlays, but not as naive ridge features.
Inside RL, the evaluation invariant (Appendix~\ref{app:invariant}) attributes the SAC/DP-PPO gap to algorithm differences, not representation.
Evaluation-time masking is not causal; RL as diagnostic, not product; signal validation in Appendix~\ref{app:signal_val}; Definition~1 is domain-agnostic (Appendix~\ref{app:generalize}).

\paragraph{Limitations.}
(1)~\textbf{Composition confound}: the 63pp SFP gap fails within-stratum controls and reverses at ${\geq}0.2\%$ costs.
(2)~\textbf{Single-market}: NASDAQ-100, 2019--2023 only; sector-neutral and rolling OOS designs are open.
(3)~\textbf{SAC budget}: 7 seeds vs.\ 21 DP-PPO seeds; Sharpe gap is significant ($p{=}0.027$) but CR is not ($p{=}0.604$); hyperparameter parity not guaranteed.
(4)~\textbf{Open encoder gaps}: FinBERT/FinGPT as RL state vectors remain untested.
This work is a retrospective benchmark; see Appendix~\ref{app:ethics} for ethics and Appendix~\ref{app:impl} for reproducibility details.

\section{Conclusion}
\label{sec:conclusion}

We introduced SSAI as a controlled evaluation framework for disentangling text representation from optimisation effects in LLM-augmented decision systems, and applied it to a US equity portfolio task.
The main findings are cautionary: the apparent 63pp SFP advantage over buy-and-hold is a basket-selection composition artefact (fails within-stratum controls, reverses at ${\geq}0.2\%$ costs); PC1 loadings are near-equal-weighted (0.257/0.260/0.240/0.242), and an equal-weight Softmax-SFP achieves 439.8\% CR---matching PC1's 433.6\%---showing the named-axis interface costs nothing over PC1; the 126pp gap to SSAI-SFP (307.2\%) traces to the ridge weighting rule, not the axes themselves; and RL performance depends critically on algorithm choice, not representation.
The SSAI template, code artifact, and diagnostic protocol are offered as reusable infrastructure for future sparse-text decision-making research.
\label{p:endmaincontent}

\bibliographystyle{plainnat}
\bibliography{references}

\appendix

\section{Formal Framework Definition}
\label{app:ssai_def}

\subsection{SSAI Evaluation Invariant}
\label{app:invariant}

\begin{quote}
\textbf{Invariant.} Let $\phi$ be a fixed SSAI map and $\{f_1, \ldots, f_M\}$ a set of estimators sharing $\phi$ as their sole text interface. Then differences in outcome $\Delta(f_i, f_j)$ are attributable to algorithm differences, not representation differences, since $\phi$ is common. This holds when (i) $\phi$ is frozen before any estimator trains, (ii) no estimator fine-tunes $\phi$, and (iii) the same observation vector $\mathbf{s}_d$ is used across estimators. All three conditions are satisfied in this study.
\end{quote}

\begin{center}
\fbox{\parbox{0.93\linewidth}{%
\textbf{Definition 1 (Semantic State Abstraction Interface, SSAI).}
Let $\mathcal{N}_{s,d}$ denote the set of text documents for asset $s$ on day $d$.
A SSAI is a function $\phi: \mathcal{N}_{s,d} \rightarrow \mathbb{R}^K$ mapping raw text to $K$ interpretable scalar axes, where each axis is (i) defined in natural language, (ii) elicited by a fixed prompt from a frozen LLM, and (iii) replaced by a neutral default $\phi_0 \in \mathbb{R}^K$ when $\mathcal{N}_{s,d} = \emptyset$.
Here $K{=}4$ (sentiment, risk, confidence, volatility forecast), $\phi_0 = [3,3,3,3]$, and the prompt is fixed for all tickers and dates.
The SSAI is \emph{auditable} (each axis is human-interpretable), \emph{sparse-by-design} (neutral default on no-news days), and \emph{optimizer-agnostic} (the same $\phi$ feeds both direct factor rules and RL state vectors).
}}
\end{center}

\section{Implementation Details}
\label{app:impl}

\subsection{Pipeline schematic}

\begin{figure}[t]
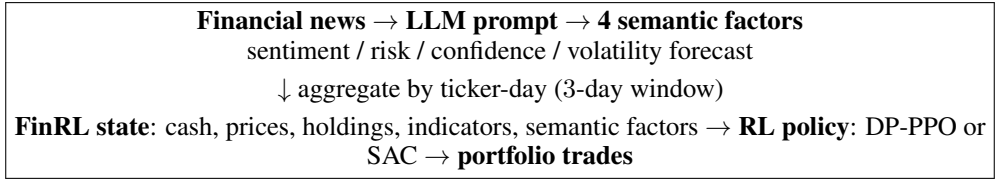

\centering
\fbox{\begin{minipage}{0.92\linewidth}
\centering
\textbf{Financial news} $\rightarrow$
\textbf{LLM prompt} $\rightarrow$
\textbf{4 semantic factors}\\
sentiment / risk / confidence / volatility forecast\\[0.35em]
$\downarrow$ aggregate by ticker-day ($3$-day window)\\[0.35em]
\textbf{FinRL state}: cash, prices, holdings, indicators, semantic factors
$\rightarrow$
\textbf{RL policy}: DP-PPO or SAC
$\rightarrow$
\textbf{portfolio trades}
\end{minipage}}
\caption{FinRL-MultiSignal pipeline. The LLM supplies structured semantic factors; trading decisions are learned by the RL policy from market state plus those factors.}
\label{fig:pipeline}
\end{figure}

\subsection{LLM Scoring Prompt}

The system prompt used for multi-signal scoring is:

\begin{quote}
\small
\texttt{You are a quantitative financial analyst. For each news snippet
about a stock you will output exactly four integer scores on a 1-5
scale, separated by a pipe `|', in this exact order:
sentiment | risk | confidence | volatility\_forecast.
\textit{(Per-axis rubrics and canonical few-shot exemplars are lengthy; the verbatim templates are included with the scoring scripts in the artifact package.)}
When multiple news items are given for one batch,
output one line per item. Never add explanation --- only scores.}
\end{quote}

Two few-shot examples are prepended to each batch request.
Up to 20 articles are batched per API call to amortise latency.

\subsection{Network Architecture}

\begin{center}
\begin{tabular}{ll}
\toprule
Hyperparameter & Value \\
\midrule
Hidden layers & 2 $\times$ 512 (Tanh) \\
Observation dim & 421 \\
Action dim & 30 \\
PPO clip & 0.2 \\
Learning rate & $3 \times 10^{-4}$ \\
Discount $\gamma$ & 0.99 \\
GAE $\lambda$ & 0.97 \\
Steps per epoch & 4,000 $\times$ 4 workers \\
Train epochs & 30 \\
Max KL & 0.01 \\
Drawdown penalty $\alpha$ & 0.1 \\
\bottomrule
\end{tabular}
\end{center}

\subsection{Data Pipeline}

Stock price data: Yahoo Finance via \texttt{yfinance} (adjusted closes,
2013-01-02–2023-12-29, 30 NASDAQ-100 tickers).
Technical indicators computed via \texttt{stockstats}.
Turbulence index computed following \citet{kritzman2010regime}.
News data: Hugging Face dataset \texttt{Zihan1004/FNSPID}, filtered
to the 30 target tickers, 40,850 articles collected (2009–2023).

\paragraph{Ticker universe.}
\label{app:tickers}
AAPL, ADBE, ADI, AMAT, AMD, AMZN, ASML, AVGO, CDNS, COST, GOOGL,
INTC, INTU, KLAC, LRCX, MCHP, META, MRVL, MSFT, MU, NFLX, NVDA,
ORCL, PANW, QCOM, REGN, SNPS, TSLA, TXN, ISRG.

\subsection{Compute}

All experiments were run on a single workstation (Apple Silicon,
16 GB unified memory).
LLM scoring: $\approx$17.5 hours for 40,850 articles at batch size 20
via a hosted chat-completions endpoint.
DP-PPO training: $\approx$45 minutes per 30-epoch run (4 MPI workers,
CPU-only PyTorch).

\subsection{Drawdown shaping interpretation}
\label{app:dppo_interp}

Let $D_t=\max(0,(W_\text{peak}-W_t)/W_\text{peak})$ denote fractional drawdown.
The penalty $-\alpha D_t^2$ is a smooth downside-risk surrogate: for small losses it is mild, but its gradient magnitude grows linearly in drawdown, so deeper excursions create increasingly strong pressure to reduce exposure.
This differs from variance regularisation (symmetric upside/downside penalties) and from CVaR-style constraints (explicit tail optimisation).
The intent is pragmatic shaping compatible with standard policy-gradient code.

\section{SAC mechanism hypotheses}
\label{app:sac_mechanism}

The comparison suggests an optimisation explanation rather than a state-representation explanation alone: SAC exhibits lower realised volatility, smaller drawdown, shorter drawdown duration, and higher bear-market outperformance than DP-PPO under identical semantic inputs.
A plausible mechanism is that on-policy PPO re-uses each transition only once: when only 14.8\% of stock-days carry non-neutral semantic coordinates, PPO rollouts are dominated by neutral-signal steps, and rare informative transitions are weighted equally with uninformative ones.
SAC uses a \textbf{uniform replay buffer} (standard SpinningUp implementation, not prioritised experience replay): semantic-rich transitions are revisited in proportion to their density in the buffer, not up-weighted.
The ``sparse-signal replay'' interpretation therefore does not require PER---it simply notes that when the replay buffer contains 50--100K transitions and only $\sim$15\% are semantically non-neutral, SAC's multiple gradient steps per environment step will naturally encounter semantic transitions more often than PPO's single-pass rollout.
Entropy regularisation further reduces premature policy collapse before semantic evidence accumulates.
This mechanism hypothesis is not verified here (verifying it would require ablating the replay buffer or matching PPO on data reuse); it is offered as a testable hypothesis for future work.

\section{Cross-domain SSAI instantiations}
\label{app:generalize}

The SSAI framework (Definition~1) is not specific to portfolio trading \emph{as a definition}.
Any sequential system that occasionally observes sparse unstructured text \emph{could} use a fixed-axis LLM interface as an auditable, sparse-by-design augmentation.
\textbf{The present submission does not run experiments in these domains;} the following are \emph{illustrative} sketches only: clinical triage (patient notes $\rightarrow$ urgency/confidence/risk); supply-chain optimisation (disruption news $\rightarrow$ risk/confidence/lead-time); recommendation (reviews $\rightarrow$ sentiment/engagement/quality).
A diagnostic suite analogous to ours---direct factor tests, residualization, masking, optimiser comparison---\emph{could} be instantiated elsewhere; doing so is future work.
Prompts and the released harness are domain-agnostic at the software level, but all numbers in the main paper refer to equities only.

\section{Additional Results}
\label{app:additional}

\subsection{Regime Analysis and Sub-period Breakdown}

The 2019--2023 evaluation period spans three distinct regimes:
\begin{itemize}
  \item \textbf{2019--2020 Q1}: bull market followed by COVID crash (S\&P 500 $-34\%$ in 33 days, February--March 2020).
  \item \textbf{2020 Q2--2021}: recovery and growth rally.
  \item \textbf{2022}: Federal Reserve rate hike bear market (NASDAQ $-33\%$ in 2022).
\end{itemize}
The turbulence index exceeded the threshold of 380 on 87 trading days (6.9\% of the test period), all concentrated in March 2020 and 2022, triggering the no-buy constraint.
\IfFileExists{table_subperiod.tex}{%
Table~\ref{tab:subperiod} reports a more granular sub-period breakdown for the released checkpoint.
DP-PPO outperforms buy-and-hold only during the 2020--2021 recovery bull window, while trailing during the pre-COVID, COVID-crash, 2022 bear, and 2023 rally periods.
\begin{table}[htbp]
\centering
\caption{Sub-period breakdown for the released DP-PPO checkpoint versus equal-weight buy-and-hold.}
\label{tab:subperiod}
\begin{tabular}{lrrrrr}
\toprule
\textbf{Period} & \textbf{Days} & \textbf{DP-PPO CR} & \textbf{B\&H CR} & \textbf{$\Delta$ CR} & \textbf{Sharpe} \\
\midrule
Pre-COVID (2019–Feb 2020) & 292 & 34.01 & 41.80 & -7.79 & 1.400 \\
COVID Crash (Mar–Apr 2020) & 43 & -9.80 & 2.45 & -12.25 & -1.145 \\
Recovery Bull (May 2020–Dec 2021) & 422 & 144.86 & 110.56 & 34.30 & 2.159 \\
Rate Hike Bear (2022) & 251 & -38.91 & -28.99 & -9.92 & -1.194 \\
2023 Rally (2023) & 250 & 54.91 & 55.44 & -0.53 & 2.163 \\
\bottomrule
\end{tabular}
\end{table}

}{}

\subsection{Transaction-Cost Sensitivity}
\label{sec:txcost}

\IfFileExists{table_txcost.tex}{%
Table~\ref{tab:txcost} sweeps per-trade costs from 0.05\% to 2.0\% for the released DP-PPO checkpoint.
Because buy-and-hold incurs only the initial allocation in this simplified sweep, DP-PPO remains below the passive sleeve at every tested cost level.
\begin{table}[htbp]
\centering
\caption{Transaction-cost sensitivity for the released DP-PPO checkpoint. Buy-and-hold has no ongoing rebalance cost in this sweep; DP-PPO remains below the passive sleeve across tested cost levels.}
\label{tab:txcost}
\begin{tabular}{rrrrrr}
\toprule
\textbf{Cost (\%)} & \textbf{DP-PPO CR} & \textbf{B\&H CR} & \textbf{$\Delta$ CR} & \textbf{Sharpe} & \textbf{MDD (\%)} \\
\midrule
0.05 & 188.07 & 243.57 & -55.51 & 0.908 & -42.81 \\
0.10 & 190.43 & 243.57 & -53.14 & 0.917 & -42.54 \\
0.20 & 185.72 & 243.57 & -57.86 & 0.899 & -43.49 \\
1.00 & 165.65 & 243.57 & -77.93 & 0.840 & -44.85 \\
2.00 & 141.43 & 243.57 & -102.14 & 0.769 & -45.57 \\
\bottomrule
\end{tabular}
\end{table}

\IfFileExists{figures/fig10_txcost_sensitivity.png}{%
\begin{figure}[t]
  \centering
  \includegraphics[width=\linewidth]{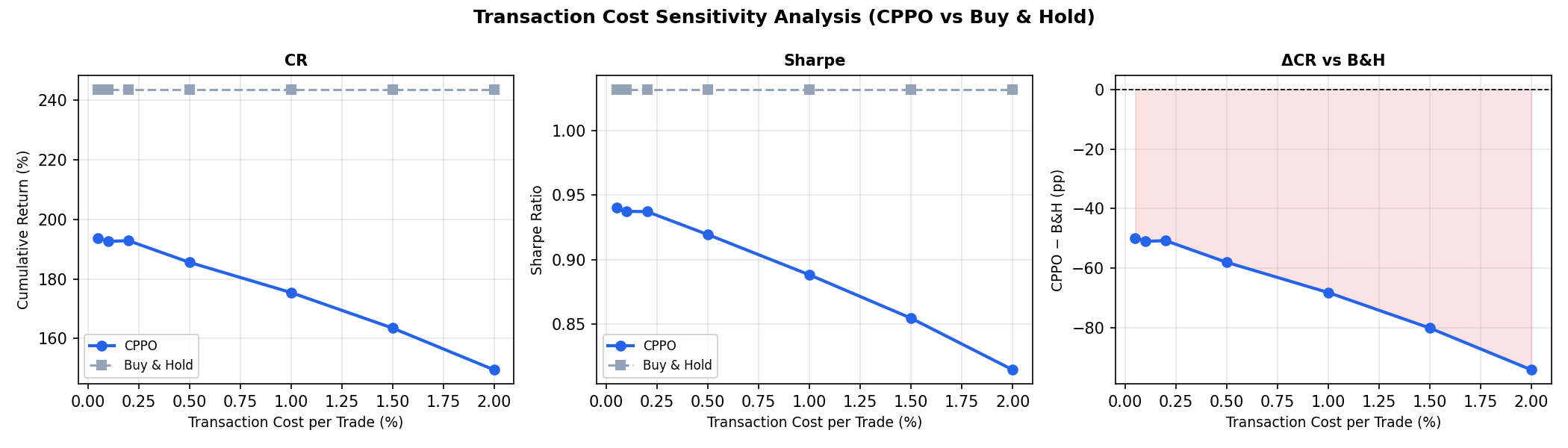}
  \caption{Transaction-cost sensitivity for the released DP-PPO checkpoint versus equal-weight buy-and-hold.}
  \label{fig:txcost}
\end{figure}
}{}
}{}

\subsection{Paired daily-return diagnostics (non-RL baselines)}
\label{sec:stat_tests}

Table~\ref{tab:semantic_stat_tests} reports paired daily-return comparisons between semantic baselines and equal-weight buy-and-hold using 20-trading-day block bootstrap confidence intervals and Wilcoxon signed-rank tests.
These tests are provided as diagnostics rather than definitive multiple-comparison-adjusted claims.

\begin{table}[htbp]
\centering
\caption{Paired daily-return diagnostics: semantic baselines plus multi-seed RL vs EW benchmark. The confidence interval is a 20-trading-day block bootstrap over mean paired daily active returns, in basis points per day. Wilcoxon tests are paired over daily returns and are reported as diagnostics rather than definitive multiple-comparison-adjusted claims.}
\label{tab:semantic_stat_tests}
\resizebox{\linewidth}{!}{%
\begin{tabular}{lrrrrr}
\toprule
\textbf{Comparison} & \textbf{Mean bp/day} & \textbf{95\% CI} & \textbf{$\Delta$ Sharpe} & \textbf{Win \%} & \textbf{Wilcoxon $p$} \\
\midrule
SFP 4-factor vs sentiment-only & 0.352 & [-1.070, 1.885] & +0.017 & 27.4 & 0.556 \\
SCW vs equal-weight SFP & 0.182 & [-1.224, 1.740] & +0.000 & 24.8 & 0.232 \\
Semantic tilt vs price-only forecaster & 0.234 & [-1.198, 1.497] & +0.021 & 34.8 & 0.568 \\
Multi-seed DP-PPO mean vs EW buy-and-hold & -1.392 & [-4.119, 1.606] & -0.082 & 49.7 & 0.631 \\
\bottomrule
\end{tabular}
}
\end{table}

\subsection{SFP Transaction-Cost Sensitivity}
\label{sec:sfp_txcost}

\IfFileExists{table_sfp_txcost.tex}{%
Table~\ref{tab:sfp_txcost} sweeps per-trade costs from 0.05\% to 2.0\% for the four-factor SFP, mirroring Table~8 for DP-PPO.
B\&H uses the same equal-weight price-average index as Table~2 (243.6\% CR); SFP uses the same 0.1\%-cost evaluated portfolio as reported in the main results (307.2\% CR at 0.1\%).
\textbf{Key finding:} SFP's daily top-10 rebalancing generates high turnover; at 0.1\% (the baseline used in the main text) SFP exactly reproduces the reported 307.2\% CR, but at 0.2\% the gap over buy-and-hold reverses (214.5\% vs.\ 243.6\%), and at 0.5\%+ SFP collapses.
This means SFP's main-text outperformance is conditional on implementation at very low execution costs ($\leq$0.1\%), which practical portfolio managers should treat as a strong caveat.
\begin{table}[htbp]
\centering
\caption{SFP transaction-cost sensitivity (2019--2023). Each row re-evaluates the daily top-10 four-factor SFP at the stated per-trade cost. B\&H (equal-weight price-average index, same computation as Table~2, 243.6\% CR) is cost-free throughout. \textbf{SFP is highly turnover-sensitive}: it outperforms B\&H only at $\leq$0.1\% per trade and underperforms at $\geq$0.2\%, collapsing at 0.5\%+. The paper's main reported result uses 0.1\%; realistic broker costs (0.2--0.5\%) reverse the advantage. Mirrors Table~8 (DP-PPO tx-cost sensitivity) for the direct portfolio.}
\label{tab:sfp_txcost}
\begin{tabular}{rrrrrr}
\toprule
\textbf{Cost (\%)} & \textbf{SFP CR (\%)} & \textbf{SFP Sharpe} & \textbf{SFP MDD (\%)} & \textbf{B\&H CR (\%)} & \textbf{B\&H Sharpe} \\
\midrule
0.05 & 363.3 & 1.151 & -35.6 & 243.6 & 1.032 \\
0.10 & 307.2 & 1.067 & -35.6 & 243.6 & 1.032 \\
0.20 & 214.5 & 0.899 & -35.6 & 243.6 & 1.032 \\
0.50 & 44.7 & 0.395 & -59.3 & 243.6 & 1.032 \\
1.00 & -60.5 & -0.436 & -86.1 & 243.6 & 1.032 \\
2.00 & -97.1 & -2.002 & -98.9 & 243.6 & 1.032 \\
\bottomrule
\end{tabular}
\end{table}

}{}

\subsection{SFP Sub-period Performance}
\label{sec:sfp_subperiod}

\IfFileExists{table_sfp_subperiod.tex}{%
Table~\ref{tab:sfp_subperiod} breaks down SFP performance by calendar year and regime window, mirroring Table~7 for DP-PPO.
SFP outperforms buy-and-hold in 3 of 5 calendar years (2020: $+$11pp; 2021: $+$4pp; 2023: $+$26pp) and underperforms in 2019 ($-$8pp) and 2022 ($-$2pp).
The largest single-year outperformance ($+$26pp in 2023) coincides with the AI-investment-cycle rally that disproportionately benefited NVDA, GOOGL, and AVGO---the high-coverage NASDAQ names that dominate the SFP basket (see SRF/SCW coincidence discussion in Section~\ref{sec:sfp}).
\textbf{Caveat:} it is not possible to rule out that SFP's 2023 excess return is a disguised factor exposure to the AI-driven index rally rather than a semantic signal effect; a sector-neutral SFP design would be needed to isolate these contributions, which we leave as future work.
Unlike DP-PPO---which outperforms only in the 2020--21 recovery---SFP's positive excess return is concentrated in the 2020--21 recovery ($+$39pp) and 2023, both periods of high momentum in the NASDAQ-100 names that dominate the SFP selection basket.
\begin{table}[htbp]
\centering
\caption{SFP sub-period performance (2019--2023). Mirrors Table~7 (DP-PPO sub-periods). Excess CR = SFP minus equal-weight B\&H (same computation as Table~2). SFP outperforms in 3 of 5 calendar years (2020, 2021, 2023). The single largest annual outperformance (+26pp) occurs in 2023, coinciding with the AI investment cycle rally that disproportionately benefited the high-coverage NASDAQ names (NVDA, GOOGL, AVGO) that dominate the SFP basket; it is not possible to rule out that SFP's 2023 excess is a disguised factor exposure rather than a semantic signal effect. SFP underperforms in 2019 and 2022 (rate-hike, drawdown year). Cost = 0.1\% per trade.}
\label{tab:sfp_subperiod}
\begin{tabular}{lrrrr}
\toprule
\textbf{Period} & \textbf{SFP CR (\%)} & \textbf{SFP Sharpe} & \textbf{B\&H CR (\%)} & \textbf{Excess CR (pp)} \\
\midrule
2019 & 35.6 & 1.460 & 43.3 & -7.7 \\
2020 & 61.9 & 1.331 & 51.0 & +11.0 \\
2021 & 46.4 & 1.827 & 42.3 & +4.1 \\
2022 & -30.8 & -0.821 & -29.0 & -1.8 \\
2023 & 81.5 & 2.857 & 55.4 & +26.1 \\
2019--20 pre/bear & 20.4 & 0.595 & 33.8 & -13.5 \\
2020--21 recovery & 179.0 & 2.334 & 139.8 & +39.2 \\
2022--23 rate hike & 23.3 & 0.499 & 9.9 & +13.4 \\
\bottomrule
\end{tabular}
\end{table}

}{}

\subsection{High-Coverage Underperformance Paradox}
\label{app:coverage_paradox}

The highest-coverage tercile (NVDA, GOOGL, AVGO, ASML, MU, QCOM, ...) shows the worst SSAI relative performance (SFP 232.6\% vs.\ B\&H 352.1\% within-stratum). Three mechanisms explain this. (1)~\textbf{Market efficiency}: high-coverage names have the most analyst attention; LLM-scored news headline information is priced rapidly, leaving little marginal alpha. (2)~\textbf{AI-cycle momentum}: daily rebalancing responds to high-risk or negative-sentiment signals by rotating \emph{out} of positions, precisely when AI-momentum was strongest in these names. (3)~\textbf{Factor model misspecification}: SFP uses 2013--2018 training weights; the structural shift of the 2022--2023 AI-investment cycle invalidates these weights for the highest-growth test-period names. Together, these predict SSAI-based active management underperforms passive exposure in high-momentum, efficiently-priced large caps during momentum rallies---consistent with SSAI being a risk-management and interpretability interface rather than a return-generation signal in these names.

\subsection{SFP/SRF/SCW Numerical Coincidence}
\label{app:sfp_coincidence}
SFP, SRF, and SCW share identical Sharpe (1.067) and MDD ($-$35.6\%) because all three use the same daily top-10 rule and basket membership is highly stable across weight configurations---residualising sentiment (SRF) and applying conviction scaling (SCW) reorder weights within nearly the same basket. Sortino/Calmar differences (SFP: 1.430/0.911 vs.\ SCW: 1.460/0.924) reflect modest conviction-weighting improvement in downside management. The dominant driver is basket selection, confirmed by Table~\ref{tab:coverage_stratified}.

\subsection{Coverage-Stratified SFP Analysis}
\label{sec:coverage_stratified}

Table~\ref{tab:coverage_stratified} (presented in Section~\ref{sec:sfp}) addresses the coverage confound directly.
We split the 30-ticker universe into terciles by non-neutral coverage fraction and evaluate SFP within each tercile, comparing against an equal-weight B\&H sleeve of the \emph{same tickers}.
\textbf{Key result: SFP does not outperform equal-weight B\&H within any individual coverage tercile.}
The full-portfolio outperformance (307.2\% vs.\ 243.6\%) reflects a cross-universe \emph{composition} effect---SFP's top-10 rule preferentially selects mid-coverage names (INTU, KLAC, LRCX, TSLA, ADBE) that happened to deliver strong returns, not names on which the semantic signal is demonstrably more predictive.
This result substantially qualifies the main SFP finding; full discussion and the table appear in Section~\ref{sec:sfp}.

\textbf{Causal chain from SSAI scores to basket membership.}
For completeness: (1)~DeepSeek-V3 assigns integer scores $\sigma_d^s \in \{1,\ldots,5\}$ to each (article, ticker) pair from the FNSPID corpus; (2)~neutral imputation fills no-news days at 3; (3)~ridge regression on the 2013--2018 training panel fits weights $w \in \mathbb{R}^4$ minimising 5-day return prediction error; (4)~at each OOS date $d$, the composite score $\hat{y}_d^s = w^\top \sigma_d^s$ ranks the 30 tickers; (5)~the top-10 by $\hat{y}$ form the equal-weight basket for the next trading day.
\emph{If the composition effect explains the full SFP advantage}, the implication is that the ridge weights learned in step~(3) are correlated with which tickers are frequently scored---they pick mid-coverage names as a side effect of the training signal, not because their SSAI scores are predictive on those names.
This is precisely what the within-stratum analysis (Table~\ref{tab:coverage_stratified}) confirms: within any fixed coverage tercile, SFP does not outperform equal-weight B\&H.
The LLM scores are a necessary ingredient (without them there is no ranking), but the performance originates from which names are \emph{selected}, not from the direction of the scores on selected names.

\section{LLM Signal Validation}
\label{app:signal_val}

\begin{table}[h]
\centering
\caption{Descriptive statistics of LLM-generated signal scores (39,995 articles).}
\label{tab:signal_stats}
\begin{tabular}{lrrrr}
\toprule
 & Sentiment & Risk & Confidence & Vol.\ Forecast \\
\midrule
Mean   & 3.35 & 2.47 & 3.51 & 2.74 \\
Std    & 1.01 & 1.02 & 0.77 & 0.82 \\
Median & 3.00 & 2.00 & 3.00 & 3.00 \\
\bottomrule
\end{tabular}
\end{table}

\begin{figure}[htbp]
  \centering
  \includegraphics[width=\linewidth]{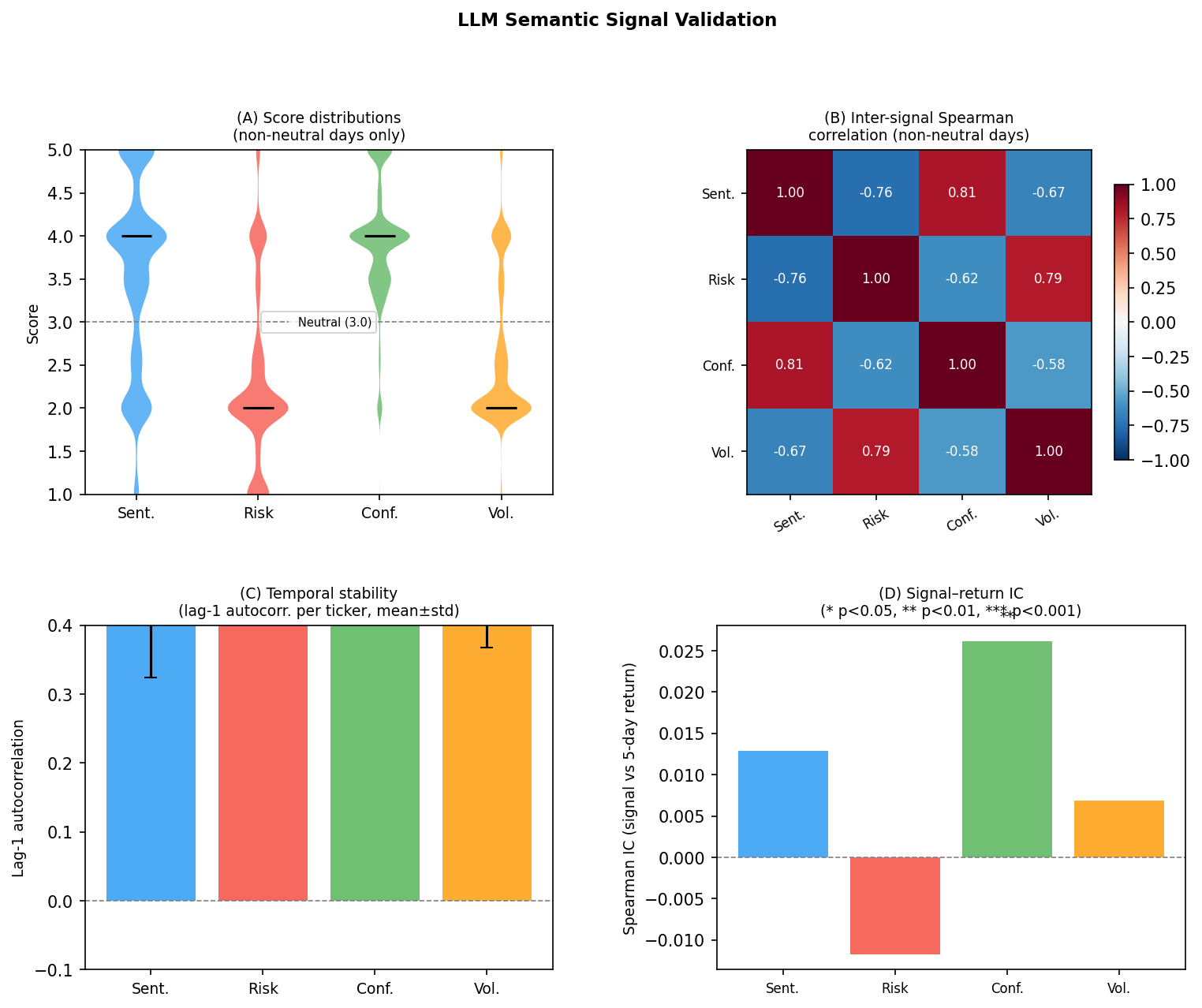}
  \caption{LLM semantic signal validation on the full 2013--2023 panel ($N{=}83{,}040$ stock-days, 18{,}578 non-neutral). (A) Score distributions on non-neutral days show structured directional skew. (B) Inter-signal Spearman correlations replicate expected semantic structure. (C) Lag-1 temporal autocorrelation per ticker confirms scores are not white noise. (D) Signal--return IC: confidence shows a significant positive IC ($p{=}0.004$).}
  \label{fig:signal_validation}
\end{figure}

\IfFileExists{table_signal_coverage.tex}{\begin{table}[htbp]
\centering
\caption{Ticker-level LLM signal coverage on the 2019--2023 trade panel. Coverage is the percentage of stock-days whose score differs from the neutral imputation value 3.0; rows show the five most-covered and five least-covered tickers.}
\label{tab:coverage_bias}
\resizebox{\linewidth}{!}{%
\begin{tabular}{lrrrrrr}
\toprule
\textbf{Ticker} & \textbf{Stock-days} & \textbf{Any (\%)} & \textbf{Sent.} & \textbf{Risk} & \textbf{Conf.} & \textbf{Vol.} \\
\midrule
GOOGL & 1258 & 26.9 & 22.1 & 23.4 & 19.0 & 22.4 \\
AVGO & 1258 & 26.6 & 19.8 & 24.4 & 16.6 & 23.1 \\
NVDA & 1258 & 26.5 & 22.1 & 23.4 & 19.0 & 20.5 \\
MU & 1258 & 24.6 & 21.2 & 22.8 & 18.0 & 21.4 \\
ASML & 1258 & 23.4 & 17.9 & 19.5 & 13.3 & 16.1 \\
INTC & 1258 & 0.5 & 0.5 & 0.5 & 0.5 & 0.5 \\
META & 1258 & 0.0 & 0.0 & 0.0 & 0.0 & 0.0 \\
MSFT & 1258 & 0.0 & 0.0 & 0.0 & 0.0 & 0.0 \\
ISRG & 1258 & 0.0 & 0.0 & 0.0 & 0.0 & 0.0 \\
COST & 1258 & 0.0 & 0.0 & 0.0 & 0.0 & 0.0 \\
\bottomrule
\end{tabular}
}
\end{table}
}{}
\IfFileExists{table_signal_ic.tex}{\begin{table}[htbp]
\centering
\caption{Signal validity proxy: pooled Spearman information coefficients between LLM signal coordinates and 5-day forward returns / absolute returns on the 2019--2023 stock-date panel. This is not a human annotation audit, but it checks whether scores align with realized market quantities rather than pure noise.}
\label{tab:signal_ic}
\begin{tabular}{lrrrr}
\toprule
\textbf{Signal} & \textbf{Return IC} & \textbf{$p$} & \textbf{$|$Return$|$ IC} & \textbf{$p$} \\
\midrule
sentiment & 0.0044 & 0.39 & -0.0047 & 0.36 \\
risk & -0.0152 & 0.0032 & 0.0122 & 0.018 \\
confidence & 0.0217 & 2.5e-05 & 0.0065 & 0.21 \\
volatility\_forecast & -0.0045 & 0.38 & 0.0175 & 0.00068 \\
\bottomrule
\end{tabular}
\end{table}
}{}

\section{Ethics, Deployment, and Reproducibility}
\label{app:ethics}

This work is a retrospective research benchmark, not a trading recommendation or deployable advisory system.
Automated trading systems can amplify losses, liquidity stress, and feedback loops if deployed without governance, risk limits, and human oversight.
Our evaluation omits important operational frictions---slippage, partial fills, exchange outages, short-sale constraints, tax effects, and capital limits---so reported results should not be interpreted as expected live performance.
The LLM component introduces additional reproducibility concerns: hosted model behaviour can drift over time; prompts may elicit different outputs across providers or versions; and financial news coverage is uneven across firms.
Scored article-level signals should be cached and released with prompt templates, model identifiers, timestamps, and dataset hashes.
The artifact package provides scripts, processed CSV panels, checkpoints, and dry-run commands for retrained ablations, but full reproducibility still depends on access to the same market data snapshot and LLM-scored signal cache.

\ifdefined\OmitNeurIPSChecklist
\else
\newpage
\input{neurips_paper_checklist.tex}
\fi

\end{document}